\documentclass{article}

\usepackage{microtype}
\usepackage{graphicx}
\usepackage{subfigure}
\usepackage{booktabs} 

\usepackage{hyperref}

\usepackage{microtype}
\usepackage{graphicx}
\usepackage{subfigure}
\usepackage{booktabs} 
\usepackage{algorithm}
\usepackage{algorithmic}
\usepackage{multirow}
\usepackage{xcolor}
\usepackage{amsmath}
\usepackage{tabularx}
\usepackage{enumitem}
\usepackage{bm}
\usepackage[utf8]{inputenc} 
\usepackage[T1]{fontenc}    
\usepackage{hyperref}       
\usepackage{booktabs}       
\usepackage{amsfonts}       
\usepackage{nicefrac}       
\usepackage{microtype}      
\usepackage{amsthm}
\usepackage{xr}
\usepackage{authblk}
\usepackage{amssymb}

\usepackage[toc,page]{appendix}

\newtheorem{theorem}{Theorem}
\newtheorem{lemma}[theorem]{Lemma}

\newtheorem{corollary}[theorem]{Corollary}
\newtheorem{definition}[theorem]{Definition}
\newtheorem{problem}[theorem]{Problem}

\usepackage[accepted]{icml2021}

\icmltitlerunning{Efficient Optimal Transport Algorithm by Accelerated Gradient descent}

\begin{document}

\twocolumn[
\icmltitle{Efficient Optimal Transport Algorithm by Accelerated Gradient descent}




\begin{icmlauthorlist}
\icmlauthor{Dongsheng An}{sbu}
\icmlauthor{Na Lei}{dut}
\icmlauthor{Xianfeng Gu}{sbu}
\end{icmlauthorlist}

\icmlaffiliation{sbu}{Department of Computer Science, Stony Brook University, NY, USA}
\icmlaffiliation{dut}{DUT-RU ISE, Dalian University of Technology, Liaoning, China}

\icmlcorrespondingauthor{Dongsheng An}{doan@cs.stonybrook.edu}
\icmlcorrespondingauthor{Na Lei}{nalei@dlut.edu.cn}
\icmlcorrespondingauthor{Xianfeng Gu}{gu@cs.stonybrook.edu}


\vskip 0.3in
]



\printAffiliationsAndNotice{}  

\begin{abstract}
Optimal transport (OT) plays an essential role in various areas like machine learning and deep learning.
However, computing discrete optimal transport plan for large scale problems with adequate accuracy and efficiency is still highly challenging. 
Recently, methods based on the Sinkhorn algorithm add an entropy regularizer to the prime problem and get a trade off between efficiency and accuracy. 
In this paper, we propose a novel algorithm to further improve the efficiency and accuracy based on Nesterov's smoothing technique. 
Basically, the non-smooth c-transform of the Kantorovich potential is approximated by the smooth Log-Sum-Exp function, which finally smooths the original non-smooth Kantorovich dual functional. The smooth Kantorovich functional can be optimized by the fast proximal gradient algorithm (FISTA) efficiently. Theoretically, the computational complexity of the proposed method is given by $O(n^{\frac{5}{2}} \sqrt{\log n} /\epsilon)$, which is lower than 
current estimation of the Sinkhorn algorithm.
Empirically, compared with the Sinkhorn algorithm, our experimental results demonstrate that the proposed method achieves faster convergence and better accuracy with the same parameter.
\end{abstract}

\vspace{-3mm}
\section{Introduction}
Optimal transport (OT) is a powerful tool to compute the Wasserstein distance between probability measures, which are widely used to model various natural and social phenomena, including economics \cite{Galichon2016economy}, optics \cite{Glimm2003optical}, biology \cite{Schiebinger2019cell},  physics  \cite{Jordan1998differential} and so on. Recently, optimal transport has been successfully applied in the areas of machine learning and statistics, such as parameter estimation in Bayesian nonparametric models \cite{Nguyen2013Bayesian}, computer vision \cite{arjovsky2017wgan, Courty2017domain, An2019AEOT, An2020AEOTGAN}, natural language processing \cite{Kusner2015wordembedding, Yurochkin2019hot} and so on. In these areas, the complex probability measures are approximated by summations of Dirac measures supported on the  samples. To obtain the Wasserstein distance between the empirical distributions, we have to solve the discrete OT problem.

For the discrete optimal transport problem, where both the source and target measures are discrete, the Kantorovich functional becomes a convex function defined on a convex domain. Due to the lack of smoothness, conventional gradient descend method can not be applied directly. Instead, it can be optimized with the sub-differential method \cite{Nesterov2005nonsmooth}, where the gradient is replaced by the sub-differential. In order to achieve an approximation error less than $\varepsilon$, the sub-differential method requires $O(1/\varepsilon^2)$ iterations. Recently, several approximation methods have been proposed to improve the computational efficiency. In these methods \cite{Cuturi2013Sinkhorn, Benamou2015sinkhorn, Altschuler2017greenhorn}, a strongly convex entropy function is added to the prime Kantorovich problem and thus the regularized problem can be efficiently solved by the Sinkhorn algorithm. More detailed analysis shows that the computational complexity of the Sinkhorn algorithm is $\tilde{O}(n^2/\varepsilon^2)$ \cite{Dvurechensky2018PD} by setting $\lambda=\epsilon/ 4\log n$. Also, a series of primal-dual algorithms are proposed, including the APDAGD algorithm \cite{Dvurechensky2018PD} with computational complexity $\tilde{O}(n^{2.5}/\epsilon)$, the APDAMD algorithm \cite{Lin2019(APDAMD} with $\tilde{O}(n^{2}\sqrt{r}/\epsilon)$ where $r$ is a complex constant of the Bregman divergence, and the APDRCD algorithm \cite{Guo2020APDRCD} with $\tilde{O}(n^{2.5}/\epsilon)$. But all of the three methods need to build a matrix with space complexity $O(n^3)$, which makes them hard to use when $n$ is large. 

\textbf{Our Method} In this work, instead of starting from the prime Kantorovich problem like the Sinkhorn based methods, we directly deal with the dual Kantorovich problem. The key idea is to approximate the original non-smooth c-transform of the Kantorovich potential by Nesterov's smoothing idea. Specifically, we approximate the $\max$ function by the Log-Sum-Exp function, which has also been used in \cite{Schmitzer2019SparseOT, peyre2018COT}, such that the original non-smooth Kantorovich functional is converted to be an unconstrained $(n-1)$-dimensional smooth convex energy. By using the Fast Proximal Gradient Method (FISTA) \cite{Beck2008FISTA}, we can quickly optimize the smoothed energy to get a precise estimate of the OT cost. In theory, the method can achieve the approximate error $\varepsilon$ with space complexity $O(n^2)$ and computational complexity ${O}(n^{2.5}\sqrt{\log n}/\varepsilon)$. 

The contributions of the proposed method includes:
    (i) We convert the dual Kantorovich problem to be an unconstrained convex optimization problem by approximating the nonsmooth c-transform of the Kantorovich potential with Nesterov's smoothing. 
    (ii) The smoothed Kantorovich functional can be efficiently solved by the FISTA algorithm with computational complexity {$\tilde{O}(n^{2.5}/{\sqrt{\varepsilon}})$}. 
    At the same time, the computational complexity of the Kantorovich functional itself 
    is given by $\tilde{O}(n^{2.5}/\varepsilon)$. 
    (iii) The experiments demonstrate that compared with the Sinkhorn algorithm, the proposed method achieves faster convergence, better accuracy and higher stability with the same parameter $\lambda$.

\section{Optimal Transport Theory}
\label{sec:theory}
In this section, we introduce some basic concepts and theorems in the classical OT theory, focusing on Kantorovich's approach and its generalization to the discrete settings via c-transform. The details can be found in Villani's book \cite{villani2008optimal}.

\begin{problem}[Kantorovich Problem] 
Suppose $X\subset \mathbb{R}^d$, $Y\subset \mathbb{R}^d$  are two subsets of the Euclidean space $\mathbb{R}^d$, $\mu=\sum_{i=1}^m \mu_i \delta(x-x_i), \nu=\sum_{j=1}^n \nu_j \delta(y-y_j)$ are two discrete probability measures defined on $X$ and $Y$ with equal total measure, $\sum_{i=1}^m \mu_i=\sum_{j=1}^n \nu_j$.
Then the Kantorovich problem aims to find the optimal transport plan $P=(p_{ij})$ that minimizes the total transport cost:
\begin{equation}
\setlength{\abovedisplayskip}{3pt}
\small
(KP)\hspace{2mm}  M_c(\mu,\nu) = \min_{P \in \pi(\mu,\nu)} \sum_{i=1}^m \sum_{j=1}^n p_{ij}c_{ij}.
    \label{eq:KP}
\setlength{\belowdisplayskip}{3pt}
\end{equation}
where $c_{ij}=c(x_i, y_j)$ is the cost that transports one unit mass from $x_i$ to $y_j$, and $\pi(\mu,\nu) = \{P|\sum_{i=1}^m p_{ij}=\nu_j, \sum_{j=1}^n p_{ij} = \mu_i, p_{ij}\geq 0\}$.
\end{problem}

\begin{problem}[Dual Kantorovich] Given two discrete probability measures $\mu=\sum_{i=1}^m \mu_i \delta(x-x_i), \nu=\sum_{j=1}^n \nu_j \delta(y-y_j)$ supported on $X$ and $Y$, and the transport cost function $c: X\times Y\to \mathbb{R}$, the Kantorovich problem is equivalent to maximizing the following \emph{\bf Kantorovich functional}:
\begin{equation}
\begin{split}
\small
(DP1)~ M_c(\mu,\nu) =    \max_{\phi, \psi}\left\{
    -\sum_{i=1}^m \phi_i \mu_i + \sum_{j=1}^n \psi_j \nu_j
    \right\}    
\end{split}
    \label{eq:DP1}
\end{equation}
where $\phi\in L^1(X,\mu)$ and $\psi\in L^1(Y,\nu)$ are called the \emph{\bf Kantorovich potentials}, and $-\phi_i + \psi_j \le c_{ij}$. The above problem can be reformulated as the following minimization form with the same constraints:
\begin{equation}
\small
(DP2)~M_c(\mu,\nu)=    -\min\left\{
    \sum_{i=1}^m \phi_i \mu_i - \sum_{j=1}^n \psi_j \nu_j
    \right\}
    \label{eq:DP2}
\end{equation}
\end{problem}

\begin{definition}[c-transform]
Let $\phi\in L^1(X,\mu)$ and $\psi\in L^1(Y,\nu)$, we define 
\begin{equation}
\small
    \phi_i = \psi^c(x_i) = \max_j \{\psi_j-c_{ij}\}  
    \label{eqn:discrete_c_transform}
\end{equation}
\end{definition}

With c-transform, Eqn. (\ref{eq:DP2}) is equivalent to solving the following unconstrained convex optimization problem:
\begin{equation}
\small
    M_c(\mu,\nu) = -\min_{\psi} E(\psi) = -\min_{\psi} \{\sum_{i=1}^m \mu_i\psi^c(x_i) - \sum_{j=1}^n \nu_j \psi_j\}
    \label{eq:cpt2}
\end{equation}
Suppose $\psi^*$ is the solution to the problem Eqn.~(\ref{eq:cpt2}), then $\psi^* + k\mathbf{1}$ is also an optimal solution for Eqn.~(\ref{eq:cpt2}).
In order to make the solution unique, we add a constraint $\psi\in H$ using the indicator function $I_H$, where $H=\{\psi|\sum_{j=1}^n \psi_j = 0\}$, and modify the \textbf{Kantorovich functional} $E(\psi)$ in Eqn.~(\ref{eq:cpt2}): 
\begin{equation}
\small
\tilde{E}(\psi) = E(\psi) + I_H(\psi), \quad I_H(\psi) = \left\{
\begin{array}{ll}
0& \psi \in H\\
\infty& \psi \not\in H
\end{array}
\right.
\label{eqn:indicator_energy}
\end{equation}
Then the problem Eqn.~(\ref{eq:cpt2}) is equivalent to solving:
\begin{equation}
\small
\begin{split}
    M_c(\mu,\nu) &= -\min_\psi \tilde{E}(\psi) \\
    &=-\min_\psi \sum_{i=1}^m \mu_i\psi^c(x_i)-\sum_{j=1}^n \nu_j\psi_j + I_H(\psi)
\end{split}
    \label{eq:cp3}
\end{equation}
which is essentially an $(n-1)$-dimensional unconstrained convex problem. According to the definition of $c$-transform in Eqn.~(\ref{eqn:discrete_c_transform}), $\psi^c$ is non-smooth w.r.t $\psi$.

\section{Nesterov's Smoothing of Kantorovich functional}


In this section, we smooth the non-smooth discrete Kantorovich functional $E(\psi)$ by approximating $\psi^c(x)$ with the Log-Sum-Exp function to get the smooth Kantorovich functional $E_\lambda(\psi)$, following Nesterov's original strategy \cite{Nesterov2005nonsmooth}, which has also been applied in the OT field \cite{peyre2018COT, Schmitzer2019SparseOT}. Then through the FISTA algorithm \cite{Beck2008FISTA}, we can easily induce that the computation complexity of our algorithm is $O(n^{2.5}\sqrt{\log n}/\varepsilon)$, with $\tilde{E}(\psi^t)-\tilde{E}(\psi^*) \leq \epsilon$. By abuse of notation, in the following we call both $E(\psi)$ and $\tilde{E}(\psi)$ the Kantorovich functional, both $E_\lambda(\psi)$ and $\tilde{E}_\lambda(\psi)$ the smooth Kantorovich functional. 

\begin{definition}[$(\alpha, \beta)$-smoothable]
A convex function $f$ is called $(\alpha,\beta)$-smoothable if for any $\lambda>0$, $\exists$ a convex function $f_\lambda$ s.t.
\begin{equation*}
\small
    \begin{split}
        f_\lambda(x) &\leq f(x) \leq f_\lambda(x) + \beta \lambda \\
        f_\lambda(y) &\leq f_\lambda(x) + \langle \nabla f_\lambda(x), y-x \rangle + \frac{\alpha}{2\lambda}(y-x)^T \nabla^2f_\lambda(x)(y-x)
    \end{split}
\end{equation*}
Here $f_\lambda$ is called a $\frac{1}{\lambda}$-smooth approximation of $f$ with parameters $(\alpha, \beta)$.
\end{definition}
In the above definition, the parameter $\lambda$ defines a tradeoff between the approximation accuracy and the smoothness, the smaller the $\lambda$, the better approximation and the less smoothness.

\begin{lemma}[Nesterov's Smoothing]
Given $f:\mathbb{R}^n\to\mathbb{R}$, $f(x) = \max\{x_j:j=1,\dots,n\}$, for any $\lambda>0$, we have its $\frac{1}{\lambda}$-smooth approximation with parameters $(1,\log n)$
\begin{equation}
\small
    f_\lambda(x) = \lambda \log \left( \sum_{j=1}^n e^{{x_j}/{\lambda}} \right) - \lambda \log n,
    \label{eqn:f_lambda}
\end{equation}
\end{lemma}

Recall the definition of c-transform of the Kantorovich potential in Eqn.~(\ref{eqn:discrete_c_transform}), we obtain the Nesterov's smoothing of $\psi^c$ by applying Eqn.~(\ref{eqn:f_lambda})
\begin{equation}
\small
\psi^c_\lambda(x_i) = \lambda \log \left(
\sum_{j=1}^n e^{(\psi_j-c_{ij})/\lambda}
\right) - \lambda \log n.
\label{eqn:smoothed_Kantorovich_potential}    
\end{equation}
We use $\psi_\lambda^c$ to replace $\psi^c$ in Eqn.~(\ref{eq:cp3}) to approximate the Kantorovich functional.
Then the Nesterov's smoothing of the Kantorovich functional becomes
\begin{equation}
\small
    E_\lambda(\psi) = \lambda \sum_{i=1}^m \mu_i \log \left(\sum_{j=1}^n e^{(\psi_j-c_{ij})/\lambda}\right) - \sum_{j=1}^n \nu_j \psi_j -\lambda \log n
    \label{eqn:dskp}
\end{equation}
and its gradient is given by 
\begin{equation}
\small
\frac{\partial E_\lambda (\psi)}{\partial \psi_j} = \sum_{i=1}^m \mu_i \frac{e^{(\psi_j-c_{ij})/\lambda}}{\sum_{k=1}^n e^{(\psi_k-c_{ik})/\lambda}} - \nu_j, ~~\forall ~j\in [n]
\label{eqn:discrete_E_lambda_grad}
\end{equation}
Furthermore, we can directly compute the Hessian matrix of $E_\lambda(\psi)$. Let $K_{ij} = e^{-c_{ij}/\lambda}$ and $v_j=e^{\psi_j/\lambda}$, and set $ E_\lambda^i:= \lambda \log \sum_{j=1}^n K_{ij}v_j,~\forall ~i\in[m]$.
Direct computation gives the following Hessian matrix:
\begin{equation}
\small
\begin{split}
    \nabla^2 E_\lambda &= \frac{1}{\lambda}\sum_{i=1}^m\mu_i  \left(\frac{1}{\mathbf{1}^TV_i} \Lambda_i - \frac{1}{(\mathbf{1}^TV_i)^2} V_i V_i^T\right) 
\end{split}
\label{eqn:Hessian}
\end{equation}
where $V_i=(K_{i1}v_1,K_{i2}v_2,\dots,K_{in}v_n)^T$, and $\Lambda_i = \text{diag}(K_{i1}v_1,K_{i2}v_2,\dots,K_{in}v_n)$. By the Hessian matrix, we can show that $E_\lambda$ is a smooth approximation of $E$.


\begin{lemma} $E_\lambda(\psi)$ is a $\frac{1}{\lambda}$-smooth approximation of $E(\psi)$ with parameters $(1,\log n)$.
\label{lem:nesterov}
\end{lemma}

\begin{lemma} Suppose $E_\lambda(\psi)$ is the $\frac{1}{\lambda}$-smooth approximation of $E(\psi)$ with parameters $(1,\log n)$, $\psi_\lambda^*$ is the optimizer of $E_\lambda(\psi)$, the approximate OT plan is unique and given by 
\begin{equation}
\small
\begin{split}
    (P_\lambda^*)_{ij} = \mu_i \frac{e^{((\psi_\lambda^*)_j-c_{ij})/\lambda}}{\sum_{k=1}^n e^{((\psi_\lambda^*)_k-c_{ik})/\lambda}} 
    = \frac{\mu_i K_{ij}v_j^*}{K_iv^*}
    \label{eq:ot_plan}
    \end{split}
\end{equation}
where $K_i$ is the $ith$ row of $K$ and $v^*=e^{\psi_\lambda^*/\lambda}$.
\end{lemma}
Similar to the discrete Kantorovich functional in Eqn.~(\ref{eq:cpt2}), the optimizer of the smooth Kantorovich functional in Eqn.~(\ref{eqn:dskp}) is also not unique: given an optimizer $\psi_\lambda^*$, then $\psi_\lambda^*+k\mathbf{1}$, $k\in\mathbb{R}$ is also an optimizer. We can eliminate the ambiguity by adding the indicator function as Eqn.~(\ref{eqn:indicator_energy}), $\Tilde{E}_\lambda(\psi) = E_\lambda(\psi) + I_H(\psi)$,
\begin{equation}
\small
\begin{split}
    \Tilde{E}_\lambda(\psi) &= \lambda \sum_{i=1}^m \mu_i \log \left(\sum_{j=1}^n e^{(\psi_j-c_{ij})/\lambda}\right) 
    - \sum_{j=1}^n \nu_j \psi_j -\lambda \log n + I_H(\psi)
\end{split}
\label{eqn: smooth_energy}
\end{equation}
This energy can be optimized effectively through the accelerated proximal algorithms \cite{Nesterov2005nonsmooth}, such as the 
Fast Proximal Gradient Method (FISTA) \cite{Beck2008FISTA}. The FISTA iterations are as follows:
\begin{equation}
\small
    \begin{split}
        z^{t+1} &= \Pi_{\eta_t I_H}\left(\psi^t-\eta_t\nabla E_\lambda(\psi^t)\right)\\
        \psi^{t+1} &= z^{t+1} + \frac{\theta_t-1}{\theta_{t+1}}(z^{t+1}-z^t) 
    \end{split}
    \label{eq:update}
\end{equation}
with initial conditions $\psi^0=v^0=\mathbf{0}$, $\theta_0=1$, $\eta_t=\lambda$ and $\theta_{t+1}=\frac{1}{2}\left(1+\sqrt{1+4\theta_t^2}\right)$. Here $\Pi_{\eta_t I_H}(z)=z-\frac{1}{n}\sum_{j=1}^n z_j$ is the projection of $z$ to $H$ (the proximal function of $I_H(x)$) \cite{Parikh2014proximal}. Similar to the Sinkhorn's algorithm, this algorithm can be parallelized, since all the operations are row based.


\begin{theorem}
Given the cost matrix $C=(c_{ij})$, the source measure $\mu$ and target measure $\nu$ with $\sum_{i=1}^m \mu_i = \sum_{j=1}^n \nu_j = 1$, $\psi^*$ is the optimizer of the discrete dual Kantorovich functional $\tilde{E}(\psi)$, and $\psi_\lambda^*$ is the optimizer of the smooth Kantorovich functional $\tilde{E}_\lambda(\psi)$. Then the approximation error is 
\begin{equation*}
\small
    |\tilde{E}(\psi^*) - \tilde{E}_\lambda(\psi_\lambda^*)| \leq 2\lambda\log n
\end{equation*}

\end{theorem}

\begin{corollary}
Suppose $\lambda$ is fixed and $\psi^0=\mathbf{0}$, then for any 
$t\geq \sqrt{\frac{{2\|\psi_\lambda^*\|^2}}{\lambda \varepsilon}}$, we have 
\begin{equation*}
\small
\tilde{E}_\lambda(\psi^t) - \tilde{E}_\lambda(\psi_\lambda^*) \leq \varepsilon, 
\end{equation*}

where $\psi_\lambda^*$ is the optimizer of $\tilde{E}_\lambda(\psi)$.
\end{corollary}

\begin{theorem}
If $\lambda=\frac{\varepsilon}{2\log n}$, 
then for any $t \geq \frac{\sqrt{8\|\bar{C}\|^2 n\log n}}{\epsilon}$ with $\bar{C}=C_{\max}-\lambda\log \nu_{min}$, we have
\begin{equation*}
\small
   \tilde{E} (\psi^t)- \tilde{E}(\psi^{*}) < \varepsilon
\end{equation*}
where $\psi^t$ is the solver of $\tilde{E}_\lambda(\psi)$ after $t$ steps in the iterations in Eqn. (\ref{eq:update}), and $\psi^*$ is the optimizer of $\tilde{E}(\psi)$. Then the total computational complexity is $O(\frac{n^{2.5}\sqrt{ \log n}}{\varepsilon})$.
\label{thm:convergence}
\end{theorem}
Since $\tilde{E} (\psi^t) > \tilde{E} (\psi_\lambda^*)$, thus we have 
\begin{equation}
\small
   0<\tilde{E} (\psi_\lambda^*) - \tilde{E}(\psi^{*}) < 2\lambda\log n
   \label{eq:linear_convergence}
\end{equation}

This shows that $\tilde{E} (\psi_\lambda^*)$ at least linearly converges to $\tilde{E}(\psi^{*})$ with respect to $\lambda$.
\begin{figure*}[h]
\centering
\begin{tabularx}{\linewidth}{cccc}
    \includegraphics[width=0.23\linewidth]{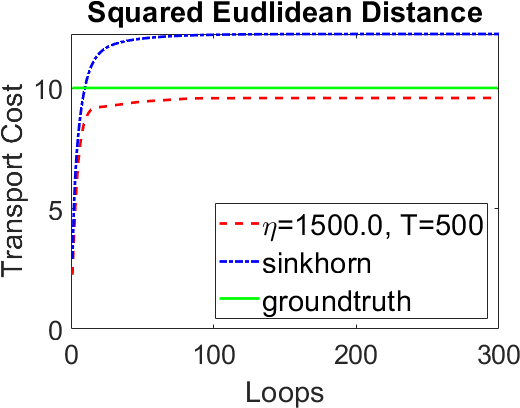} &
    \includegraphics[width=0.23\linewidth]{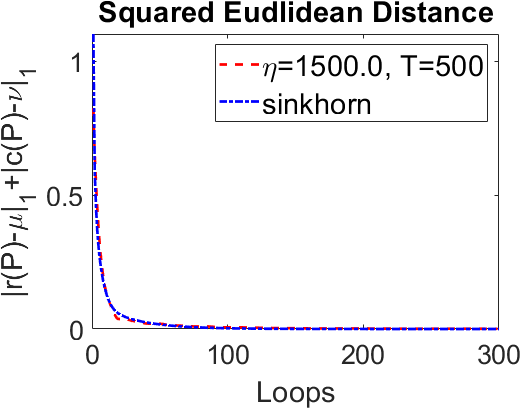} &
    \includegraphics[width=0.23\linewidth]{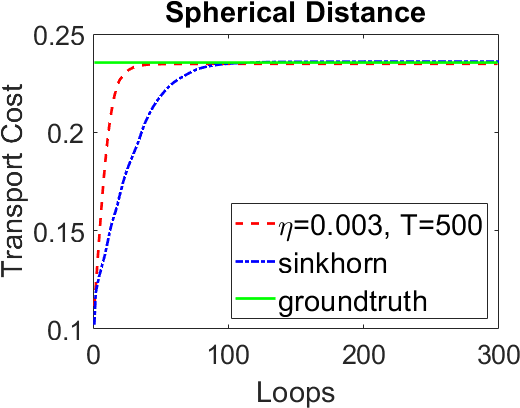} &
    \includegraphics[width=0.23\linewidth]{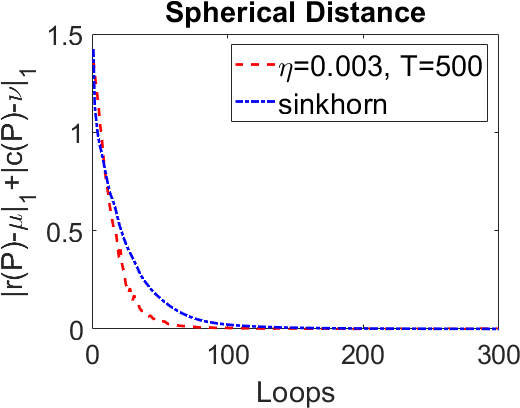} 
\end{tabularx}
\vspace{-3mm}
\caption{Comparison with the Sinkhorn algorithm \cite{Cuturi2013Sinkhorn} under different costs.}
    \label{fig:compare}
    \vspace{-6mm}
\end{figure*}

\section{Experiments}
In this section, we investigate the performance of the proposed algorithm by comparing with the Sinkhorn algorithm \cite{Cuturi2013Sinkhorn} with computational complexity $\tilde{O}(n^2/\epsilon^2)$ \cite{Dvurechensky2018PD}. 
All of the codes are written using MATLAB with GPU acceleration. The  experiments are conducted on a Windows laptop with Intel Core i7-7700HQ CPU, 16 GB memory and NVIDIA GTX 1060Ti GPU.

\textbf{Cost Matrix} We test the performance of the algorithm with different cost matrix. Specifically, we set $\mu=\sum_{i=1}^m \mu_i \delta(x-x_i)$, $\nu=\sum_{j=1}^n \nu_j \delta(y-y_j)$. After the settings of $\mu_i$s and $\nu_j$s, they are normalized by $\mu_i = \frac{\mu_i}{\sum_{k=1}^m \mu_k}$, $\nu_j = \frac{\nu_j}{\sum_{k=1}^n \nu_k}$. To build the cost matrix, we use the squared Euclidean distance and the spherical distance.
\begin{itemize}
    \item For the squared Euclidean distance (SED) experiment, like \cite{Altschuler2017greenhorn}, we randomly choose one pair of images from the MNIST dataset \cite{mnist2010}, and then add negligible noise $0.01$ to each background pixel with intensity 0. $\mu_i$ and $x_i$ ($\nu_j$ and $y_j$) are set to be the value and the coordinate of each pixel in the source (target) image. Then the squared Euclidean distance between $x_i$ and $y_j$ are given by  $c(x_i,y_j)=\|x_i-y_j\|^2$.
    \item For the spherical distance (SD) experiment, we set $m=n=500$. Both $\mu_i$s and $\nu_j$s are randomly generated from the uniform distribution $Uni([0,1])$. $x_i$'s are randomly sampled from the Gaussian distribution $\mathbb{N}(3\mathbf{1_d}, I_d)$ and $y_j$'s are randomly sampled from the Uniform distribution $Uni([0,1]^d)$. Then we normalize $x_i$ and $y_j$ by $x_i=\frac{x_i}{\|x_i\|_2}$ and $y_j = \frac{y_j}{\|y_j\|_2}$. As a result, both $x_i$'s and $y_j$'s are located on the sphere. The spherical distance is given by $c(x_i, y_j) = \arccos{(\langle x_i, y_j\rangle)}$.
\end{itemize}

\textbf{Evaluation Metrics} W mainly use two metrics to evaluate the proposed method: the first is the total transport cost, which is defined by Eqn.~(\ref{eq:cpt2}) and given by $-E(\psi)$; and the second is the $L_1$ distance from the computed transport plan $P_\lambda$ to the admissible distribution space $\pi(\mu, \nu)$ defined as $D(P_\lambda)=\|P_\lambda\mathbf{1}-\mu\|_1+\|P_\lambda^T\mathbf{1}-\nu\|_1$.




Then we compare with the Sinkhorn algorithm \cite{Cuturi2013Sinkhorn} w.r.t both the convergence rate and the approximate accuracy. We manually set $\lambda= \frac{\max(C)-\min(C)}{T}$ with $T=500$ to get a good estimate of the OT cost, and adjust the step length $\eta$ to get the best convergence of the proposed method. For fair comparison, we use the same $\lambda$ for the Sinkhorn algorithm. We summarize the results in Fig.~\ref{fig:compare}, where the green curves represent the groundtruth computed by linear programming, the blue curves are for the Sinkhorn algorithm, and the red curves give the results of our method. 

From the figure, we can observe that
in both experiments, our method achieves faster convergence than the Sinkhorn algorithm. 
Also, $-E(\psi_\lambda^*)$ gives a better approximate of the OT cost than $\langle P_\lambda, C \rangle$ used in the Sinkhorn algorithm with the same small $\lambda$, especially for the squared euclidean distance cost. Basically, to achieve $\epsilon$ precision, $\langle P_\lambda^*, C \rangle$ needs to set $\lambda = \frac{\epsilon}{4\log n}$ \cite{Dvurechensky2018PD}, which is smaller than our requirement for $\lambda=\frac{\epsilon}{2\log n}$.

We also report the iterations and the corresponding running time with $T=700$ in Tab. \ref{tab:compare}. The stop condition is set to be $|E(\psi^{t+1})-E(\psi^{t})|/|E(\psi^{t})|<10^{-3}$. In each iteration, Sinkhorn requires $2n^2$ times of operation, and the proposed method requires $3n^2$ times of operation. From Tab. \ref{tab:compare} we can see that the proposed method needs less iterations and time for convergence.
\begin{table}[]
\centering
\small
\caption{The running iterations and time comparison between the proposed method and Sinkhorn \cite{Cuturi2013Sinkhorn} with $T=700$.}
\label{tab:compare}
\begin{tabular}{ccccc}
\hline
\multirow{2}{*}{} & \multicolumn{2}{c}{Sinkhorn} & \multicolumn{2}{c}{Ours} \\
\cline{2-5}
                  & Iterations     & Time(s)     & Iterations   & Time(s)   \\
\hline
SED               & 209            & 0.0431      & 29          & 0.0197        \\
\hline
SD                & 297             & 0.0564      & 22       & 0.0142   \\
\hline 
\end{tabular}
\vspace{-7mm}
\end{table}
\section{Conclusion}
In this paper, we propose a novel algorithm to improve the accuracy for solving the discrete OT problem based on Nesterov's smoothing technique.
The c-transform of the Kantorovich potential is approximated by the smooth Log-Sum-Exp function, and the smoothed Kantorovich functional can be solved by the fast proximal gradient algorithm (FISTA) efficiently. Theoretically, the computational complexity of the proposed method is given by $O(n^{2.5} \sqrt{\log n} /\varepsilon)$, which is lower than current estimation of the Sinkhorn method. Empirically, our experimental results demonstrate that the proposed method achieves faster convergence and better accuracy than the Sinkhorn algorithm.


\nocite{langley00}

\bibliography{example_paper}
\bibliographystyle{icml2021}

\newpage
\appendix
\section{The Proofs}
In this section, we give the proofs of the lemmas, theorems and propositions in the main paper.

\begin{proof}[\textbf{Proof of Lemma 5}]
We have $\forall x\in \mathbb{R}^n$, 
\begin{equation*}
    \begin{split}
        f_\lambda(x) &\leq \lambda \log(n \max_j e^{x_j/\lambda})-\lambda \log(n) = f(x)  \\
        f(x) &= \lambda\log \max_j e^{x_j/\lambda} < \lambda\log \sum_{j=1}^n e^{x_j/\lambda} = f_\lambda(x)+\lambda \log n
    \end{split}
    \label{eq:log-sum-exp}
\end{equation*}
Furthermore, it is easy to prove that $f_\lambda (x)$ is $\frac{1}{\lambda}$-smooth. Therefore, $f_\lambda(x)$ is an approximation of $f(x)$ with parameters $(1, \log n)$.
\end{proof}



\begin{proof}[\textbf{Proof of Lemma 6}]
From Eqn.~(\ref{eqn:Hessian}), we see $\nabla^2 E_\lambda^i= \frac{1}{\lambda}\left(\frac{1}{\mathbf{1}^TV_i} \Lambda_i - \frac{1}{(\mathbf{1}^TV_i)^2} V_i V_i^T\right) $ has $O=\{k\mathbf{1}:k\in\mathbb{R}\}$ as its null space. In the orthogonal complementary space of $O$, $\nabla^2 E_\lambda^i$ is diagonal dominant, therefore strictly positive definite. 

Weyl's inequality \cite{Horn1991matrix} states that the eigen value of $A=B+C$ is no greater than the maximal eigenvalue of $B$ minus the minimal eigenvalue of $C$, where $B$ is an exact matrix and $C$ is a perturbation matrix. Hence the maximal eigenvalue of $\nabla^2 E_\lambda^i$, denoted as $\lambda_i$, has an upper bound,
\[
0\le \lambda_i \le \frac{1}{\lambda} \frac{1}{\mathbf{1}^TV_i} \max_j \{K_{ij}v_j\}  \le \frac{1}{\lambda}.
\]

Thus the maximal eigenvalue of $\nabla^2E_\lambda(\psi)$ is no greater than $\sum_{i=1}^n \mu_i\lambda_i =\frac{1}{\lambda}$.
It is easy to find that $E_\lambda(\psi) \leq E(\psi) \leq E_\lambda(\psi) + \lambda \log n$. Thus,$E_\lambda (\psi)$ is a $\frac{1}{\lambda}$-smooth approximation of $E(\psi)$ with parameters $(1, \log n)$.
\end{proof}

\begin{proof}[\textbf{Proof of Lemma 7}] By the gradient formula Eqn.~(\ref{eqn:discrete_E_lambda_grad}) and the optimizer $\psi_\lambda^*$, we have 
\[
\frac{\partial E_\lambda (\psi_\lambda^*)}{\partial \psi_j} = \sum_{i=1}^m (P_\lambda^*)_{ij} - \nu_j = 0 \quad \forall ~j=1,\cdots,n.
\]
On the other hand, by the definition of $P_\lambda^*$, we have
\[
    \sum_{j=1}^n (P_\lambda^*)_{ij} = \mu_i, \quad \forall ~i=1,\dots,m,
\]
compare the above two equations, we obtain that $P_\lambda^* \in \pi(\mu, \nu)$ is the approximate OT plan.

The optimal solvers of $E_\lambda(\psi)$ has the form $\psi_\lambda^*+\mathbf{1}$, all of them induce the same approximate transport plan given by Eqn. (\ref{eq:ot_plan}). Thus, $P_\lambda^*$ is unique.

\end{proof}

\begin{proof}[\textbf{Proof of Theorem 8}]
Assume $\psi^*$ and $\psi_\lambda^*$ are the minimizers of $E(\psi)$ and $E_\lambda(\psi)$ respectively. Then by the inequality in Eqn.~(\ref{eq:log-sum-exp})
\[
\begin{array}{llll}
E_\lambda(\psi^*) &\le E(\psi^*) &\le E(\psi_\lambda^*) &\le E_\lambda(\psi_\lambda^*) + \lambda \log n\\
E_\lambda(\psi_\lambda^*) &\le E_\lambda(\psi^*) &\le E(\psi^*) &\le E_\lambda(\psi^*) + \lambda \log n
\end{array}
\]
This shows $|E_\lambda(\psi^*)-E_\lambda(\psi_\lambda^*)|\le \lambda\log n$. Removing the indicator functions, we can get
\[
\begin{split}
|\tilde{E}(\psi^*) - \tilde{E}_\lambda(\psi_\lambda^*)| &= |E(\psi^*) - E_\lambda(\psi_\lambda^*)|\\ 
    &\le |E(\psi^*) - E_\lambda(\psi^*) | + |E_\lambda(\psi^*) -  E_\lambda(\psi_\lambda^*)| \\
    & \leq 2\lambda \log n.
\end{split}
\]
\end{proof}

\begin{proof}[\textbf{Proof of Corollary 9}]
Under the settings of the smoothed Kantorovich problem ~Eqn.~(\ref{eqn:dskp}),  $E_\lambda(\psi)$ is convex and differentiable with $\nabla^2 E_\lambda(\psi) \preceq \frac{1}{\lambda} I $, $I_H(\psi)$ is convex and its proximal function is given by $\Pi_H(v)$. Thus, directly applying Thm 4.4 of \cite{Beck2008FISTA} and set $L=\frac{1}{\lambda}$, we can get $\tilde{E}_\lambda(\psi^t)-\tilde{E}_\lambda(\psi_\lambda^*)\leq \frac{2}{\lambda (t+1)^2}\|\psi_\lambda^*-\psi^0\|^2$. Set $\psi^0 = \textbf{0}$ and $\frac{2}{\lambda (t+1)^2}\|\psi_\lambda^*\|^2 \leq \varepsilon$, then we get that when $t\geq \sqrt{\frac{2\|\psi_\lambda^*\|^2}{\lambda\varepsilon}}$, we have $\tilde{E}_\lambda(\psi^t)-\tilde{E}_\lambda(\psi_\lambda^*) \leq \varepsilon$.
\end{proof}

\begin{proof}[\textbf{Proof of Theorem 10}]
We set the initial condition $\psi^0 = \mathbf{0}$. For any given $\varepsilon>0$, we choose iteration step $t$, such that $\frac{2}{\lambda (t+1)^2}\|\psi_\lambda^*\|^2\le \frac{\varepsilon}{2}$, $t\geq \frac{\sqrt{8 \|\psi_\lambda^*\|^2 \log n}}{\epsilon}$, where $\psi_\lambda^*$ is the optimizer of $\tilde{E}_\lambda(\psi)$. 
By Thm 4.4 of \cite{Beck2008FISTA}, we have 
\begin{equation*}
    \begin{split}
        \tilde{E}_\lambda(\psi^t) - \tilde{E}(\psi^*)
        & \leq \tilde{E}_\lambda(\psi^t) -\tilde{E}_\lambda(\psi^*) \\
        & \leq \tilde{E}_\lambda(\psi^t) - \tilde{E}_\lambda(\psi_\lambda^*)\\
        &\leq \frac{2}{\lambda (t+1)^2}\|\psi_\lambda^*\|^2 \\
        &\le \frac{\varepsilon}{2}
    \end{split}
\end{equation*}

By Eqn.~(\ref{eqn: smooth_energy}), we have
\begin{equation*}
    \begin{split}
        \tilde{E} (\psi^t)- \tilde{E} (\psi^{*}) &= E(\psi^t) + I_H(\psi^t) - E(\psi^*) - I_H(\psi^*) \\
        &= (E(\psi^t) - E_\lambda(\psi^t)) + (E_\lambda(\psi^t) + I_H(\psi^t)) \\
        &~~~~- (E(\psi^*) + I_H(\psi^*)) \\
        &\leq |E(\psi^t) - E_\lambda(\psi^t)| + (\tilde{E}_\lambda(\psi^t) - \tilde{E}(\psi^*))\\
        &\le \lambda \log n + \frac{\varepsilon}{2} = \frac{\varepsilon}{2} + \frac{\varepsilon}{2} \\
        &= \varepsilon
    \end{split}
\end{equation*}

Next we show that $\|\psi_\lambda^*\|^2 \leq n\|\bar{C}\|^2$ by proving $|(\psi_{\lambda}^*)_j| \leq \bar{C}~\forall j\in [n]$.
According to Eq. (\ref{eq:ot_plan}), 
\begin{equation}
    \begin{split}
        \nu_j &=\sum_{i=1}^m \mu_i \frac{e^{((\psi_\lambda^*)_j-c_{ij})/\lambda}}{\sum_{k=1}^n e^{((\psi_\lambda^*)_k-c_{ik})/\lambda}} \\
        &= \sum_{i=1}^m \mu_i \frac{e^{(\psi_\lambda^*)_j/\lambda}}{\sum_{k=1}^n e^{(\psi_\lambda^*)_k/\lambda}e^{(c_{ij}-c_{ik})/\lambda}}
    \end{split}
\end{equation}
Assume $(\psi_\lambda^*)_{\max}$ is the maximal element of $\psi_\lambda^*$, we have $\sum_{k=1}^n e^{(\psi_\lambda^*)_k/\lambda}e^{(c_{ij}-c_{ik})/\lambda} \geq e^{(\psi_\lambda^*)_{\max}/\lambda}e^{-C_{\max}/\lambda}$, where $C_{\max}$ is the maximal element of the matrix of $C$. Thus,
\begin{equation*}
    \begin{split}
        \nu_j &\leq \sum_{i=1}^m \mu_i \frac{e^{(\psi_\lambda^*)_j/\lambda}}{e^{(\psi_\lambda^*)_{\max}/\lambda}e^{-C_{\max}/\lambda}} \\
        &= ~~~~~~~~\frac{e^{(\psi_\lambda^*)_j/\lambda}}{e^{(\psi_\lambda^*)_{\max}/\lambda}e^{-C_{\max}/\lambda}}
    \end{split}
\end{equation*}
Then, $(\psi_\lambda^*)_{\max} \leq (\psi_\lambda^*)_{j} +C_{\max}-\lambda\log \nu_j$ and 
\begin{equation}
    \begin{split}
        (\psi_\lambda^*)_{\max} & \leq \frac{1}{n} \sum_{j=1}^n \{ (\psi_\lambda^*)_{j} +C_{\max}-\lambda\log \nu_j \} \\
        &\leq C_{\max} - \lambda \log \nu_{min} 
    \end{split}
    \label{eq:proof_upper_bound}
\end{equation}

According to the inequality of arithmetic and geometric means
, we have $\sum_{k=1}^n e^{(\psi_\lambda^*)_k/\lambda} \geq ne^{\frac{1}{n}(\sum_{k=1}^n (\psi_\lambda^*)_k/\lambda)}=n$. Thus, $\nu_j \leq \frac{e^{(\psi_\lambda^*)_j/\lambda}}{ne^{-C_{\max}/\lambda}}$.

\begin{equation}
    \begin{split}
        (\psi_\lambda^*)_j &\geq \lambda \log n - C_{\max} + \lambda \log \nu_j \\
        & \geq \lambda \log \nu_{min} - C_{\max}
    \end{split}
    \label{eq:proof_lower_bound}
\end{equation}
Combine Eqn. (\ref{eq:proof_upper_bound}) and (\ref{eq:proof_lower_bound}) together, we have $|(\psi_\lambda^*)_j| \leq C_{\max} - \lambda \log \nu_{min} = \bar{C}$. 
Hence, we obtain when $t\geq \frac{\sqrt{8 \|\bar{C}\|^2 n\log n}}{\epsilon}$,  $\tilde{E} (\psi^t)- \tilde{E}(\psi^{*}) < \varepsilon$.

For each iteration in Eqn. ~(\ref{eq:update}), we need $O(n^2)$ times of operations, thus total the computational complexity of the proposed method is $O(\frac{n^{2.5}\sqrt{ \log n}}{\varepsilon})$.
\end{proof}



\end{document}